\documentclass[a4paper, 10pt, conference]{IEEEtran}
\IEEEoverridecommandlockouts
\usepackage{cite}
\usepackage{amsmath,amssymb,amsfonts}
\usepackage{algorithmic}
\usepackage{graphicx}
\usepackage{textcomp}
\usepackage{xcolor}
\usepackage{multirow}
\def\BibTeX{{\rm B\kern-.05em{\sc i\kern-.025em b}\kern-.08em
		T\kern-.1667em\lower.7ex\hbox{E}\kern-.125emX}}

\begin{document}
	
	\title{Enhanced Vehicle Re-identification for ITS: A Feature Fusion approach using Deep Learning\\
	}
	\makeatletter
	\newcommand{\linebreakand}{%
	\end{@IEEEauthorhalign}
	\hfill\mbox{}\par
	\mbox{}\hfill\begin{@IEEEauthorhalign}
	}
	\makeatother
	
	\author{\IEEEauthorblockN{Ashutosh Holla B}
		\IEEEauthorblockA{\textit{Department of Information and Communication Technology} \\
			\textit{Manipal Institute of Technology,} \\ 
			\textit{Manipal Academy of Higher Education, Manipal, India }\\
			\textit{ashutoshholla28@gmail.com}}
		\and
		\IEEEauthorblockN{Manohara Pai M.M}
			\IEEEauthorblockA{\textit{Department of Information and Communication Technology} \\
			\textit{Manipal Institute of Technology,} \\ 
			\textit{Manipal Academy of Higher Education, Manipal, India }\\
			\textit{mmm.pai@manipal.edu}
		}
		\linebreakand 
		\IEEEauthorblockN{Ujjwal Verma}
		\	\IEEEauthorblockA{\textit{Department of Electronics and Communication Engineering} \\
			\textit{Manipal Institute of Technology Bengaluru} \\ 
			\textit{Manipal Academy of Higher Education, India }\\
			\textit{ujjwal.verma@manipal.edu}
		}
		\and
		\IEEEauthorblockN{Radhika M. Pai}
			\IEEEauthorblockA{\textit{Department of Information and Communication Technology} \\
			\textit{Manipal Institute of Technology,} \\ 
			\textit{Manipal Academy of Higher Education, Manipal, India }\\
			\textit{radhika.pai@manipal.edu}}}

	\maketitle
	
	\begin{abstract}
In recent years, the development of robust Intelligent transportation systems (ITS) is tackled across the globe to provide better traffic efficiency by reducing frequent traffic problems. As an application of ITS, vehicle re-identification has gained ample interest in the domain of computer vision and robotics. Convolutional neural network (CNN) based methods are developed to perform vehicle re-identification to address key challenges such as occlusion, illumination change, scale, etc. The advancement of transformers in computer vision has opened an opportunity to explore the re-identification process further to enhance performance. In this paper, a framework is developed to perform the re-identification of vehicles across CCTV cameras. To perform re-identification, the proposed framework fuses the vehicle representation learned using a CNN and a transformer model. The framework is tested on a dataset that contains 81 unique vehicle identities observed across 20 CCTV cameras. From the experiments, the fused vehicle re-identification framework yields an mAP of 61.73$\%$ which is significantly better when compared with the standalone CNN or transformer model.  
	\end{abstract}
	\begin{IEEEkeywords}
		CCTV, Keyframes, re-identification, CNN, Transformer, Triplet Loss  
	\end{IEEEkeywords}
	
	\section{Introduction}
	In the current era of the advanced transportation system, video surveillance plays a pivotal role in providing security and safety measures in traffic control\cite{khan2019survey}. With the increased demand across the globe for making cities “Smart cities”, investments are in place to develop robust Intelligent Transportation Systems\cite{8915694}. To fulfill this need, government authorities/Industrial representatives motivate to set up a surveillance environment to monitor the daily traffic activities. More often to monitor vehicle movements, surveillance cameras have been set up across prominent areas in the cities such as highways, junction points, gated campuses, etc. where there is a possibility of traffic breach. These surveillance videos are used as a source of evidence to enforce a penalty on the owner of vehicles who have violated traffic rules or caused accidents. The data acquired by surveillance cameras are valuable in performing computer vision tasks such as object counting, object detection\cite{8627998}, semantic segmentation\cite{9563599}\cite{Girisha20214115} object re-identification\cite{9022301}\cite{ni2020adaptive}\cite{Chu_2019_ICCV}, and tracking\cite{Wu_2021_CVPR}. Over the years, re-identification has made its mark in various tasks such as crowd monitoring and anomaly detection. As an application of ITS, vehicle re-identification has attracted various researchers in the field of computer vision. Vehicle re-identification aims in obtaining a possible match of a vehicle observed in one camera with images of the same vehicle appearing in the initial camera or different non-overlapping cameras \cite{Shankar_2019_CVPR_Workshops}. Compared to person re-identification, vehicle re-identification poses several challenges namely (a) there are limited appearance cues apart from vehicle color to differentiate vehicles of the same make and color (b) Identical vehicles appearing at multiple cameras are subjected to different viewpoints which lead to falsely re-identifying the vehicle of interest.
\par Several Convolutional Neural Network (CNN)-based algorithms were designed specifically to address the vehicle re-identification problems. These methods utilize several convolutional layers, pooling layers, and multiple linear layers along with stacked non-linear activation functions. However, CNN models developed for re-identification focuses on discriminative regions due to the choice of the different receptive field. During the process of convolving an image, the CNN models downsample the spatial resolution of output feature maps at different convolutional layers. Due to this, the network fails to identify similar-looking vehicles that differ with minor appearance changes \cite{he2021transreid}. To explore long-range dependencies, attention-based CNN models were introduced that aim to identify long-range dependencies \cite{lian2022transformer}. These methods utilize vehicle key points which are trained in a supervised fashion to learn the discriminative parts of the vehicle. The performance of these methods is determined by the selection of key points that are either obtained by manual annotation or automated detection.
\par With the increasing popularity of transformers in the domain of NLP, researchers have utilized the transformer models to address computer vision-related tasks. By utilizing the concept of multi-head attention, the transformers aim to capture long-range dependencies by attending to various vehicle parts compared to CNN models \cite{han2020survey}. In contrast to CNN models, Transformer models process the images at patch levels across several layers without downsampling the images which enables them to learn more local information about vehicles. Although transformer models are proven to outscore existing CNN architectures, these models require a larger dataset to yield a comparable score.
\par Inspired by these observations, in this study, a vehicle re-identification framework is proposed that performs re-identification using the vehicle features that are computed by a CNN and a transformer model. Specifically, a ResNetmid CNN model\cite{yu2017devil} and a Swin Transformer\cite{liu2021swin} are used to learn the vehicle representations of vehicles observed across nonoverlapping cameras. For a given query vehicle, its presence is verified across gallery images by generating the features learned by both ResNetmid and Swin transformer. The generated features from individual models are fused to encapsulate both global and local representations of vehicles. The proposed re-identification framework is evaluated for 81 identical vehicles that are spotted across 20 CCTV cameras. To the best of our knowledge, this is the first of its kind study that utilizes both CNN and transformer models to perform the re-identification of vehicles.
\par The contribution of the paper is summarized as follows:
\begin{itemize}
    \item A first of its kind study in conducting vehicle re-identification by fusing the vehicle features of CNN and transformer model. 
    \item Performance evaluation of vehicle re-identification framework utilizing features learned using standalone CNN, transformer model, and fused feature representation (CNN+Transformer)
    
\end{itemize}

	\section{Related Work}
Computer Vision based approaches have been used for various applications such as precision agriculture \cite{UjjwalLNCS}, environmental monitoring \cite{verma2021deeprivwidth}, and surveillance \cite{holla2020efficient} to name a few. Studies on object re-identification have been mainly focused on person and vehicle entities. This section summarizes different works that are been contributed by researchers to address vehicle re-identification.

\par Recent works on object re-identification commonly utilize triplet loss as a loss metric. The authors in the work \cite{liu2016deep} proposed a two-branch deep convolutional network that projects the vehicle images to a euclidean space to measure the similarity of the two vehicles. To learn the discriminative features from deep feature embeddings, the network utilizes a Coupled-Cluster Loss (CCL).
\par The authors in \cite{kuma2019vehicle} introduced a batch sampling strategy with triplet loss to perform vehicle re-identification. The batch sample and batch weighted variants were evaluated against the standard batch hard and batch all variants of vehicle re-identification.
\par An unsupervised metric learning model \cite{antonio2018unsupervised} is developed that leverages pairwise and triplet constraints to train a re-identification model using the triplet loss similarity metric. The vehicle features are transformed from an initial input dimension into a feature space where similar identity vehicles are close together while keeping dissimilar vehicle identities far apart. A Single-shot detector is utilized to identify the vehicles appearing in a scene and is assigned to an existing or a new tracklet. The detector is built upon a VGG-16 \cite{simonyan2014very} backbone network and is trained using a COCO dataset with only vehicle class. Here the trackelts are grouped by location of videos. To compare the similarity between two vehicles, a middle frame from the tracklets is selected and the similarity score is computed using Euclidian distance.
\par A vehicle re-identification and abnormality detection framework were contributed in the work \cite{nguyen2019vehicle}. The framework constitutes three steps to obtain vehicle features at discriminative level.  The deep metric embedding module is utilized to extract discriminative vehicle features. The classifier module addresses how the vehicle features can be learned when they are of different pose and color. A Faster RCNN is used to detect the vehicles appearing on the scene. The features of detected vehicles are learned using a ResNet-50 \cite{he2016deep} trained with a triplet loss metric. As a post-optimization step, the authors have re-ranked the candidate images for a given query using the bag-of-words approach.
\par To alleviate the requirement of labeled data, the authors in \cite{wu2018vehicle} proposed an adaptive feature learning method to address re-identification. A re-identification network is trained on existing datasets by fine-tuning the feature extractor module to adapt to any different target dataset. Their proposed framework consists of three stages namely, the vehicle proposal stage, single-camera tracking, and a feature extractor step to perform re-identification.

\par As transformer architectures have found significant success in NLP, its application in computer vision tasks specific to re-identification is limited. Authors in \cite{he2021transreid} developed a transformer-based object re-identification framework that comprises two modules. Primarily these modules are designed to acquire more robust discriminative features of the vehicle and mitigate the similarity discrepancy across inter-cameras and intra-cameras matching. The authors evaluated the performance of the framework with the existing vehicle re-identification datasets.

\par Above discussed methods focus on re-identification by utilizing standard CNN architectures. With the emergence of transformer-based architectures, the problem of re-identification can be addressed significantly. There is limited set of studies that conduct re-identification utilizing both CNN and transformer-based architectures. Hence there is a scope to perform vehicle re-identification by jointly utilizing both CNN and transformer models which can overcome the hurdles faced in-vehicle re-identification such as illumination, viewpoint change, occlusion, etc.
\section{Methodology}
\begin{figure*}[!htbp]
		\begin{center}
			\includegraphics[width=\linewidth,height=7cm]{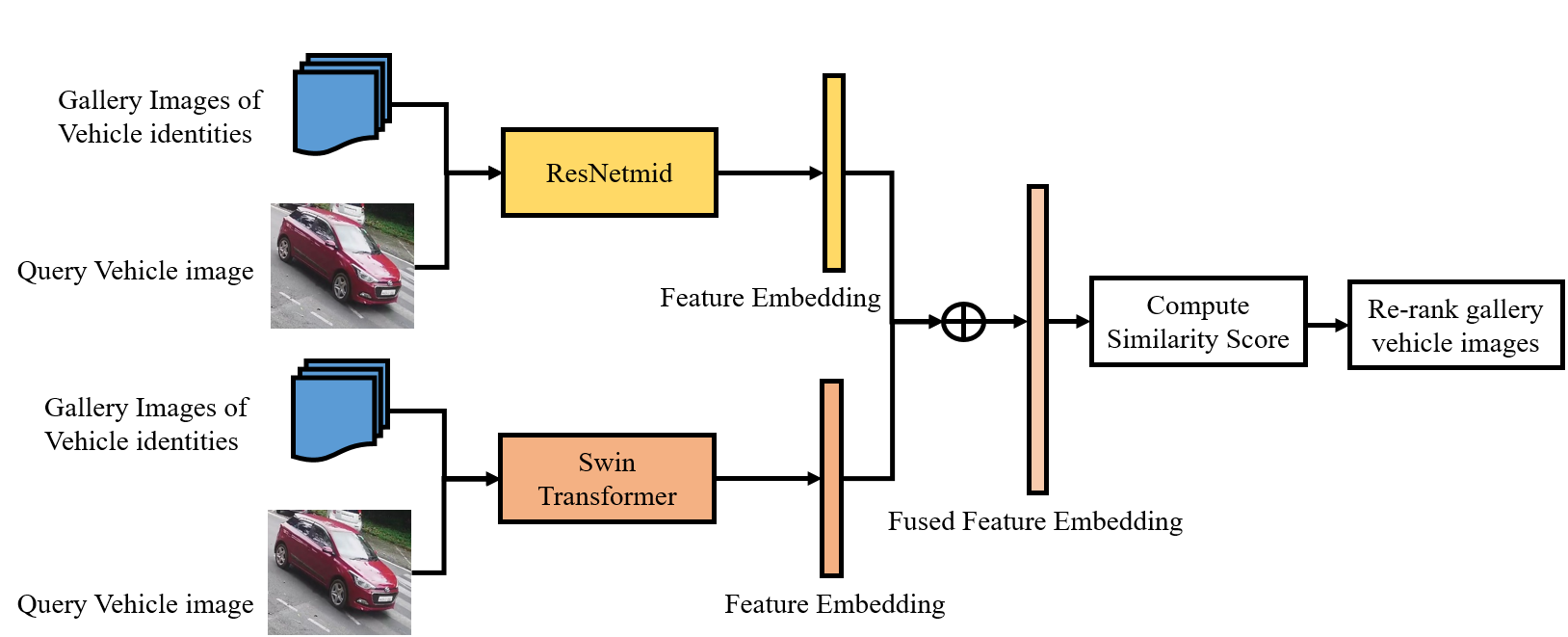}
			\caption{Outline of the proposed vehicle re-identification framework for vehicle re-identification using CCTV  surveillance systems. During inference, the network fuses the learned vehicle representations using ResNetmid and Swin transformer. A similarity score is computed for a query with entire vehicle identities appearing in the gallery. Finally, these candidate vehicle images are ranked with vehicle images appearing in gallery. }		
			\label{fig:Outline}
		\end{center}
	\end{figure*}
\subsection{Overview}
In the present work, re-identification is conducted  for vehicles observed across a network of surveillance cameras (CCTV). Vehicles observed by CCTV cameras may exhibit several appearance changes, and variations in scale, making it challenging to closely re-identify the vehicles. Hence in this work, a novel vehicle re-identification method is developed which fuses the learned feature representations from a ResNetmid\cite{he2016deep} network and Swin transformer\cite{liu2021swin}. The CNN network is utilized to learn both the semantic and global features of vehicles. The transformer network encodes the vehicle images at multiple resolutions and hierarchically fuses this learned representation at different stages. Both these networks are trained independently using triplet loss. During inference (Figure \ref{fig:Outline}), for a given query vehicle identity observed in CCTV surveillance system, its presence is verified across the gallery set that comprises identical vehicles images observed across network of cameras. This is achieved by generating the feature embeddings of vehicles using both ResNetmid and Swin transformer network. The feature embeddings are generated for query vehicle identity and vehicle identities observed by CCTV cameras. Each of the generated feature embeddings is fused and a similarity score is computed for every gallery identity vehicle and the input query vehicle. Using the similarity score the vehicle identities in the gallery are ranked such that the near resembling vehicles to the query vehicle identity appear at the top of the list. The following sections describe the overall architecture of the ResNetmid (Section \ref{subsec:resnet}) and Swin transformer (Section \ref{subsec:swin_t}) that are used to learn the vehicle representations. 
	
	\subsection{ResNetmid}
	\label{subsec:resnet}
 To learn the semantic and global features of the vehicles, in this work a ResNetmid backbone architecture is used. The architecture uses ResNet50 \cite{he2016deep} variant that comprises five residual blocks.The first five blocks of networks are initialized with pre-trained ImageNet weights. As inferred from the work\cite{yu2017devil}, utilizing the features learned from the middle layers of the network is beneficial to learning the semantic representation of the vehicles. Different from the work presented in \cite{yu2017devil}, here the final feature embeddings from residual block 4 are extracted and a global average pooling is applied. This is to encapsulate the semantic information of the vehicles that are learned by the initial residual blocks of ResNet50. The global average pooling is also applied to the feature vector generated by a final residual block of ResNet50. This feature vector exhibits a high level of the global representation of vehicles. Both the semantic feature vector and the high-level feature vector are later concatenated to generate the final feature embedding vector which incorporates both the semantic and global vehicle representation. A dense layer is added to predict the probability of a vehicle belonging to a known vehicle identity class. The network is trained using triplet loss\cite{schroff2015facenet}. The final feature embedding vector is given as the input for the loss module. Triplet loss generates triplets which consist of an anchor image a positive image that is similar to an anchor image and a negative image that is dissimilar to an anchor image.

\subsection{Swin Transformer}
	\label{subsec:swin_t}
Swin transformer processes the image at patch level by decomposing the images to several patches. Swin transformer consists of four stages. In each stage, the swin transformer generates feature maps with different sizes that correspond to different scales. The transformer initially partitions the input image into non-overlapping patches using a patch partition module. In this work a patch size of $4 \times 4$ is chosen and thus the feature dimension of each patch is $4 \times 4 \times 3=48$ . Each patch is treated as a “token” and is further projected to an arbitrary dimension ‘$C$’ using a linear embedding layer. In this work, a Swin-S variant of the transformer is utilized where the $C=96$. A swin transformer block with self-attention is applied to the tokens. At each stage, $N \times N$ neighboring patches are merged thereby performing self-attention at different scales. Different stages of swin transformer block jointly produce a hierarchical representation of the feature map of different resolutions. At each stage of the swin transformer block, the number of layers or depth of the swin transformer are {2,2,18,2}. At the last stage of the transformer, a global average pooling layer is applied to the output feature map which is later utilized to train the network with triplet loss.

\par During inference, utilizing both the networks (ResNetmid and Swin Transformer) for a given set of query vehicle images and candidate vehicle images in the gallery set, the feature embedding is generated. Specifically $ Q=\{q_{1},q_{2},....q_{N}\}$ is a set containing $N$ query vehicle images and $G = \{g_{1},g_{2},g_{3}....g_{M}\}$ denotes collection of $M$ gallery images. For each query and gallery set, the feature embeddings are obtained using the trained ResNetmid and Swin transformer model. For ResNetmid the concatenated layer that exhibits semantic and global information of vehicles is used to obtain feature embedding information for both query and gallery images. Similarly for Swin transformer, the global average pooling layer of the last stage of the transformer is used to generate the feature embeddings for both query and vehicle images present in the gallery. $F_{Resnetmid}^{Q}\in\mathbb{R}^{M\times D}$ and $F_{Swin}^{Q}\in\mathbb{R}^{M\times P}$ denotes the feature embedding representation of query set for both ResNetmid and Swin Transformer. Similarly, $F_{Resnetmid}^{G}\in\mathbb{R}^{N\times D}$ and $F_{Swin}^{G}\in\mathbb{R}^{N\times P}$ denotes the feature embedding representation of entire gallery set. Here $D$ and $P$ are the dimensions of the feature embedding layer corresponding to both ResNetmid and Swin transformer respectively. These feature representations are later fused(concatenated)  to generate a final feature vector. Utilizing these feature vectors a similarity score is computed for each query vehicle image against the vehicle images present in the gallery. , Euclidean distance is used to determine the similarity score between a query vehicle image and a candidate vehicle image appearing in the gallery.  The computed score is further sorted such that  similar appearing vehicles to the given query are closer and appear at the top in the ranked list. 
\par As illustrated in Figure \ref{fig:Outline}, during inference the generated feature embeddings for both gallery and query set of vehicles using ResNetmid and Swin Transformer are computed paralelly and are further concatenated. This process is defined in equation (\ref{eqn:eqn1}) and (\ref{eqn:eqn2})
\begin{equation}
	F_{Fused}^{Q}=Concat(F_{Resnetmid}^{Q},F_{Swin}^{Q})
	\label{eqn:eqn1}
\end{equation}
where $F_{Fused}^{Q}\in\mathbb{R}^{N \times D+P}$.
\begin{equation}
	F_{Fused}^{G}=Concat(F_{Resnetmid}^{G},F_{Swin}^{G})
	\label{eqn:eqn2}
\end{equation}
where $F_{Fused}^{G}\in\mathbb{R}^{M \times D+P}$. Both $F_{Fused}^{Q}$ and $F_{Fused}^{G}$ are utilized to compute similarity score for each of the given query image.
\section{Results and Discussion}
Here a  detailed information on the data gathering process to conduct vehicle re-identification is outlined. Further detailed analysis of vehicle re-identification using standalone ResNetmid,  Swin transformer, and the fused (concatenated) feature representation that is jointly obtained using ResNetmid and Swin transformer.
\subsection{Experimental Setup}
To evaluate the re-identification framework, surveillance data is acquired using CCTV cameras at the campus of Manipal Institute of Technology, Manipal, India. Of the entire cameras available on the campus (Total area: 188 acres), the cameras considered for this study are such that the probability of traffic movements is non-uniform. Camera locations include entry/exit of campus, academic section, hostel premises, etc. The data is collected for 2 days using Hikvision surveillance cameras with a resolution of $1920 \times 1080p$ at 20fps. A total of 81 vehicle identities were identified which are during the training and inference stage of re-identification.  The information regarding the dataset is summarized in Table \ref{tab:table1}. For a similar vehicle identified on two different days, a different vehicle identity is assigned with a belief that these vehicles may undergo appearance changes. 
\begin{table}[]
		\centering
		\caption{Description of dataset used for re-identification}
		\label{tab:table1}
		\begin{tabular}{|l|c|}
			\hline
			CCTV Cameras used for experiment          & 20 Cameras                \\ \hline
			CCTV camera frame resolution    & 1920$\times$1080p \\ \hline
			Frame rate                      & 20 fps            \\ \hline
		Duration of CCTV Videos & 15 to 40 minutes           \\ \hline
		\end{tabular}
	\end{table} 
\par Besides, processing each frame of CCTV videos is time-consuming and redundant as they are acquired at 20fps. Hence as a pre-processing step, a shot boundary detection algorithm is applied  to generate keyframes. In shot boundary detection\cite{girisha2019performance}, histogram difference is computed on an RGB image that is divided uniformly into a $16 \times 16$ non-uniform grid. A shot boundary is identified if the histogram difference between two successive frames is greater than a certain threshold value. In this paper, the threshold value is experimentally set to 0.20. The identical vehicles are manully labelled using Microsoft Vott annotation tool.
\begin{figure*}[!htbp]
		\begin{center}
			\includegraphics[width=18.5cm,height=5.5cm]{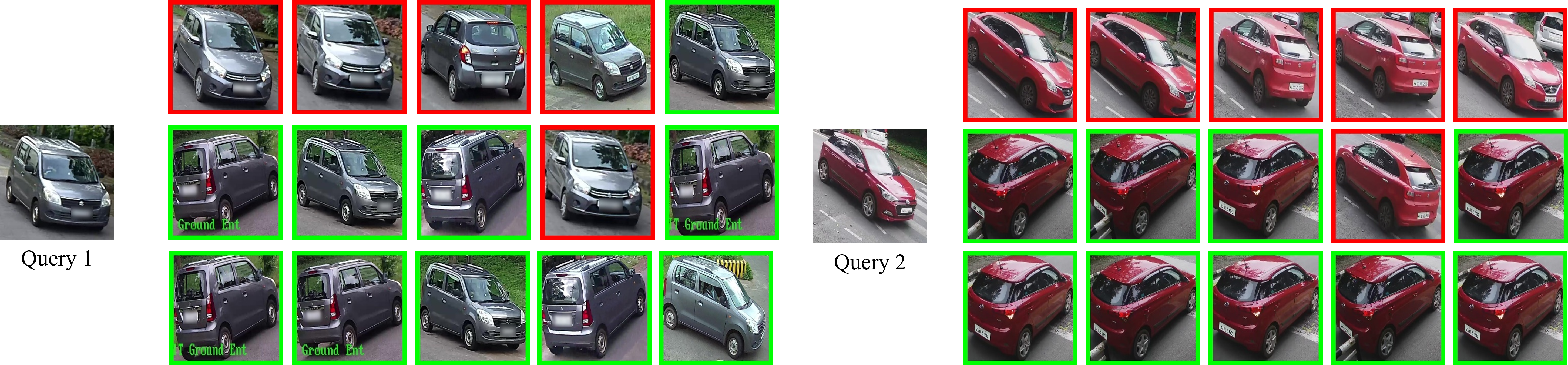}
			\caption{Illustration of top-10 results for a given query. For a given query, visualization of top-10 rank is displayed for each of the experiments i.e. ResNetmid (First row), Swin Transformer (Second row),  ResNetmid+Swin Transformer (Third row). The positive and negative match for a given query vehicle image is highlighted using green and red bounding boxes respectively. }		
			\label{fig:top-k}
		\end{center}
	\end{figure*}
\subsection{Vehicle Re-identification results}
The proposed re-identification framework utilizes features learned from both CNN(ResNetmid) and a transformer(Swin) model. Hence each network is trained individually and is further utilized during inference to obtain the vehicle representations which are later fused to compute re-identification scores. For identified 81 vehicle identities across 20 surveillance cameras, a total of 46 vehicle identities are used for training the re-identification network. A total of 1,317 annotated vehicle images of 46 vehicle identities are used to train the model. A Batch Hard triplet loss variant is used with the parameters $P$ and $K$ set to 3 and 4 respectively. The network is trained for 200 epochs with an initial learning rate of 0.001 and a decay factor of 5e-4. During inference 35 vehicle identities that were detected across 20 surveillance cameras are considered. A total of 35 images of each vehicle identity are considered as a query set. The presence of each query vehicle instance is verified across 983 vehicle images present in gallery. For each vehicle query image, a score is computed across the entire gallery of images that are further ranked with the most similar image to the given query ranked at the top of the list. 

\begin{table}[]
    \centering
    \caption{Vehicle Re-identification Results}
    \label{tab:table2}
    \begin{tabular}{|l|l|l|l|l|l|}
        \hline
        \textbf{Experiment} & \textbf{mAP}   & \textbf{rank-1} & \textbf{rank-5} & \textbf{rank-10} & \textbf{rank-20} \\ \hline
ResNetmid           & 57.78          & 71.42           & 80              & 82.85            & 85.71            \\ \hline
Swin                & 56.8           & 65.71           & 80              & 80               & 88.85            \\ \hline
ResNetmid+Swin      & \textbf{61.73} & 74.28           & 82.85           & 82.85            & 91.42            \\ \hline
\end{tabular}
\end{table}
\par The performance of the framework is assessed using mAP and rank-k accuracy. Table \ref{tab:table2} shows the re-identification scores computed for each experiment. A discussion is laid out regarding the computed mAP scores by inferring the Table \ref{tab:table2}  and the top-5 (Figure \ref{fig:top-k}) visualization obtained for each query vehicle.

\textbf{Vehicle re-identification using ResNetmid:}\\
For the developed re-identification framework, a modified architecture of ResNet50 is used to learn the semantic and global features of vehicles. When the network is solely used to infer the presence of query vehicle identities, an mAP of $57.78\%$ is obtained. From Figure \ref{fig:top-k}, for query 1 it can be observed that the network can get a single candidate match of a vehicle belonging to the same identity query vehicle image. The retrieved false match of the candidate vehicle images has a similar appearance in form of color feature. The network fails to generalize the discriminative features of vehicles thereby focussing more on the global appearance of the vehicle. For query 2 the ResNetmid fails to retrieve a candidate image similar to query vehicle identity in the top-5 rank. The retrieved top-5 candidate images have a similar appearance in form of color features. The computed rank-k scores for set of query images using ResNetmid are $71.42\%$(rank-1), $80\%$(rank-5), $82.85\%$(rank-10), $85.71\%$ (rank-20) respectively.

\textbf{Vehicle re-identification using Swin Transformer:}\\
To learn the discriminative features of the vehicle, the Swin-S variant of the transformer is used. In the network at each stage with different resolutions of the feature map, a self-attention score is computed. During inference for a given set of query vehicle images, an mAP of $56.8\%$ is obtained.
From Figure \ref{fig:top-k} it can be observed that for both the queries the network can retrieve candidate vehicle images from the gallery that are similar to the query vehicle image. Swin transformer processes the images as a collection of patches/tokens. Each patch participates in computing attention scores with neighboring patches whereby the discriminative/part-level features of vehicles are learned effectively. Hence the network can retrieve the vehicle images similar to the given query images observed at different viewpoints. The computed rank-k scores for set of query images using Swin Transformer are $65.71\%$(rank-1), $80\%$(rank-5), $80\%$(rank-10), $88.85\%$ (rank-20) respectively. 

\textbf{Vehicle re-identification using ResNetmid+Swin Transformer:}\\
As Outlined in the Figure 1, the proposed re-identification
framework fuses the feature representations of the vehicle
that are generated by two sub-networks ResNetmid and Swin
Transformer. The fused feature representation of vehicles consists of both global
features and discriminative features learned by each of the
individual network images, an mAP score of $61.73\%$ is obtained. The obtained mAP score is significantly better than the computed mAP scores obtained using the individual network. Using the fused representations, it is observed that for both query images, the network can retrieve more candidate images belonging to the query identity. Utilizing the concatenated feature representation a higher rank-k accuracy is obtained. A $74.28\%$(rank-1), $82.85\%$(rank-5), $82.85\%$(rank-10), $91.42\%$ (rank-20) scores are obtained.

\section{Conclusion}
Vehicle re-identification is open problem in computer vision task that aims to re-identify vehicles across multiple cameras.
Currently the task of performing vehicle re-identification is carried out using either CNN and attention models. These models fail to capture the long-range dependencies as the image is downsampled across deep layers whereby prominent information is lost. Transformer architectures are emerging to address computer vision tasks with greater scope to solve re-identification problems. These models process the images at the patch/token level. These patches are made to obtain the long-range dependencies across neighboring patches to weigh their importance over other patches of the image. Transformer models are computationally expensive and require a larger dataset to obtain comparable results. Under these observations, a vehicle re-identification framework is presented that fuses the learned vehicle representation from ResNetmid CNN and Swin Transformer. Using the fused vehicle representation a higher mAP of $61.73\%$ along with rank-1 of $74.28\%$ rank-5 of $82.85\%$ rank-10 of $82.85\%$ and rank-20 accuracy of $91.42\%$ is obtained. The computed scores were significantly better than re-identification scores determined individually for ResNetmid and Swin architecture. The fused representation contains both global(ResNetmid) and discriminative features(Swin Transformer) of vehicles which are useful in re-identifying vehicles that are partially occluded, subjected to the viewpoint and illumination changes.

	\small{
		\bibliographystyle{IEEEtran}
		\bibliography{Paper}
	}

\end{document}